\documentclass[twoside]{article}
\usepackage[T1]{fontenc}
\usepackage{graphicx}
\usepackage{verbatim}
\usepackage{hyperref}
\usepackage{placeins}
\usepackage{algorithm}
\usepackage{algpseudocode}
\usepackage{multirow}
\usepackage{tabularx}
\usepackage{parskip}
\usepackage{tcolorbox}
\usepackage{amssymb}
\usepackage{aligned-overset}
\usepackage{svg}
\usepackage{layouts}
\usepackage{cite}
\usepackage{amsmath,amssymb,amsfonts}
\usepackage{graphicx}
\usepackage{textcomp}
\usepackage{xcolor}
\usepackage{authblk}
\usepackage[margin=3cm]{geometry}

\graphicspath{ {./_figures/} }

\newcommand{\yplus}{\overset{+}{y}}

\newtcolorbox{mybasecolorbox}[1][]{%
  colframe=gray!25,
  title=#1
  }

\newenvironment{mytitlebox}[1][]{%
  \mybasecolorbox[#1]
  \itshape
}{%
  \endmybasecolorbox
}

\hypersetup{
     colorlinks   = true,
     citecolor    = gray
}

\usepackage{xcolor}
\usepackage{booktabs}
\usepackage{subcaption}

\usepackage{fancyhdr}
\begin{document}

\title{Beyond One-Hot-Encoding: \\Injecting Semantics to Drive Image Classifiers
}

\author[1]{Alan Perotti}
\author[1]{Simone Bertolotto}
\author[2]{Eliana Pastor}
\author[1]{André Panisson}
\affil[1]{CENTAI Institute, Turin, Italy}
\affil[2]{Politecnico di Torino, Turin, Italy}

\date{}

\maketitle

\begin{abstract}
Images are loaded with semantic information that pertains to real-world ontologies: dog breeds share mammalian similarities, food pictures are often depicted in domestic environments, and so on. However, when training machine learning models for image classification, the relative similarities amongst object classes are commonly paired with one-hot-encoded labels. According to this logic, if an image is labelled as \textit{spoon}, then \textit{tea-spoon} and \textit{shark} are equally wrong in terms of training loss.
To overcome this limitation, we explore the integration of additional goals that reflect ontological and semantic knowledge, improving model interpretability and trustworthiness. We suggest a generic approach that allows to derive an additional loss term starting from any kind of
semantic information about the classification label. 
First, we show how to apply our approach to ontologies and word embeddings, and discuss how the resulting information can drive a supervised learning process. Second, we use our semantically enriched loss to train image classifiers, and analyse the trade-offs between accuracy, mistake severity, and learned internal representations. Finally, we discuss how this approach can be further exploited in terms of explainability and adversarial robustness. 
Code repository: \href{https://github.com/S1M0N38/semantic-encodings}{https://github.com/S1M0N38/semantic-encodings}



\end{abstract}

\section{Introduction}
\label{sec:intro}
Deep Learning (DL) models have become the go-to method for addressing numerous Computer Vision (CV) tasks, such as image classification.
Unlike traditional approaches that require manual feature extraction, DL streamlines the development of end-to-end pipelines that seamlessly integrate images as inputs to the learning process, thereby automating feature extraction and enhancing overall efficiency.
This automation enables the training of DL models over extensive image datasets, which subsequently leads to enhanced model accuracy.
However, the ``black-box" nature of DL models presents challenges, as  Machine Learning (ML) practitioners often struggle to understand the chain of transformations that a DL model adopts to map an image into the final prediction. 
This lack of transparency is considered to be hampering the adoption of DL models in real-world scenarios, due to a plethora of reasons: lack of trust from domain experts, impossibility of thorough debugging from practitioners, lack of compliance to legal requirements regarding explainability, and potential systemic bias in the trained model~\cite{lipton2018mythos}. The research field of eXplainable Artificial Intelligence (XAI) tackles this problem by trying to provide more insights about the inner decision process of ML models~\cite{xai_survey}. 
However, most XAI techniques for CV are post-hoc: they are applied on trained ML models, and typically try to correlated portions of the image to the resulting label by means of input perturbation or maskings~\cite{inputxgrad,saliency,integrated_gradients}. A few other approaches try to modify the training procedure itself, hoping to gain more control over the model's internals, while at the same time maintaining competitive classification performances.

With this in mind, we remark how the standard DL pipeline for image classification trains the model to learn a mapping from images to labels. As inputs, images are loaded with semantic information that pertains to real-world ontologies: dog breeds share mammalian similarities, food
pictures are often depicted in domestic environments, and so on. As outputs, labels are typically one-hot-encoded (OHE), implementing a rigid binary logic of `one is class correct, all other classes are wrong'. There is therefore no semantics attached to these labels, as all dimensions of these OHE vectors are orthogonal. As a defective byproduct of this common ML framing, if an image is labelled as \textit{spoon}, then \textit{tea-spoon} and \textit{shark} are equally wrong predictions, and would be equally penalised in terms of training loss during a training phase. 
However, labels can often be linked them to external sources of knowledge representation, ranging from structured knowledge bases such as ontologies~\cite{wordnet} to embedding vectors~\cite{wemb} produced by language models.
Our Research Question (RQ) is therefore the following:
\begin{mytitlebox}
\textbf{RQ: How can we inject semantic information into a standard image classification learning process?}
\end{mytitlebox}
\vspace*{2mm}
In order to answer this question, in this paper we introduce a general approach that allows to inject any kind of auxiliary semantic vectors to OHE labels, and produce enriched vectors that we call \textit{Semantically-Augmented Labels} (S-AL). S-AL can be used as ground truth with standard loss functions and neural architectures in ML tasks such as image classification. We can thus formulate additional Research Sub-Questions:
\begin{mytitlebox}
\begin{itemize}
    \item \textbf{RsQ1: How does S-AL perform ML-wise, both in terms of quantity and quality of errors?}
    \item \textbf{RsQ2: What is the impact of S-AL concerning the learned internal representation of concepts?}
    \item \textbf{RsQ3: What is the impact of S-AL on XAI?}
\end{itemize}
\end{mytitlebox}

We answer all these questions empirically: first, we show how to encode both hierarchical information extracted from an ontology, and embedding vectors produced by a language model. Then we train several ML models using S-AL as targets, together with baseline models for benchmarking. We then analyse these models in terms of ML performance, structure of the feature space, and XAI heatmaps. We show how S-AL allows to maintain competitive classification accuracy while allowing for less semantically severe mistakes, and we show how the injection of semantic information improves both the space of learned features and the relative similarities of produced image explanations.

The remainder of the paper is structured as follows: Section~\ref{sec:relwork} describes related and relevant research work, while Section~\ref{sec:semloss} introduces our approach S-AL. Section~\ref{sec:exp_setup} describes how we operationalise the answer to the sub-research questions RsQ1, RsQ2, and RsQ3, and in Section~\ref{sec:exp_res} we show and discuss our experimental results. Section~\ref{sec:end} draws the  final considerations and directions for future work.

\section{Related Work}
\label{sec:relwork}
\subsection{Neural Networks for Image Classification}
Neural networks have proven to be effective in solving complex image classification tasks. Convolutional Neural Networks (CNNs) are the most widely used type of neural network for image classification. CNNs consist of multiple layers of convolutional and pooling operations that extract features from the input image. The extracted features are then fed into fully connected layers, which output the final classification probabilities.
One of the most influential works in this field was the AlexNet architecture proposed in 2012~\cite{alexnet}. The architecture used five convolutional layers and three fully connected layers to achieve state-of-the-art performance on the ImageNet~\cite{imagenet} dataset. Since then, many improvements have been made to the CNN architecture, such as the VGGNet~\cite{vgg} - however, the main structure of convolutional layers for feature extraction followed by dense layers for feature-to-label classification became a standard. A notable further development was the introduction of residual connections, starting from ResNet~\cite{resnet}. More recently, attention-based neural networks have been proposed for image classification. These models use attention mechanisms to selectively focus on important image regions, improving performance. SENet~\cite{attention} is an example of this approach, achieving state-of-the-art performance on the ImageNet by using attention modules to selectively amplify important features. 
Overall, neural networks have shown great success in image classification tasks, and their performance continues to improve with new advancements in architecture and training techniques. However, 
the image classification training process commonly involves one-hot encoding for class labels and cross-entropy as a loss function. The problem of hierarchical classification was initially explored in the literature (see survey~\cite{hier-survey}), but never incorporated into standard training pipelines.

\subsection{Semantic Auxiliary Information Sources}
Including semantic information has proved beneficial in multiple contexts, such as model interpretability~\cite{dong2017improving}, image summarization~\cite{9987488}, and image classification~\cite{mbm} itself.
However, in standard image classification tasks, labels are OHE and convey any contextual information regarding their semantic value: \textit{breakfast}, \textit{lunch} and \textit{mountain} are three equally-independent dimensions of a OHE ground truth vector. Clearly, structured representations of knowledge could provide auxiliary information concerning relations between (the concepts represented by) labels. The most common pairing of image classification dataset with external semantic information is Imagenet-Wordnet. ImageNet, as mentioned above, is a common benchmark for image classifiers; WordNet~\cite{wordnet} is a large lexical database of English. Nouns, verbs, adjectives, and adverbs are grouped into sets of cognitive synonyms (synsets), each expressing a distinct concept. Synsets are interlinked by means of conceptual-semantic and lexical relations. The already-available link between the two is that all ImageNet are also nodes (synsets) within the WordNet graph. We remark that exploiting structured representation of knowledge like WordNet is surely an interesting research direction. Still, it excludes a plethora of other unstructured semantic information sources, such as word embeddings~\cite{wemb}. A word embedding is a learned latent representation for text that allows words with similar meaning to have a similar representation and is commonly used for Natural Language Processing downstream tasks~\cite{wemb-nlp}. Typically, the representation is a real-valued vector that encodes the meaning of the word in such a way that words that are closer in the vector space are expected to be similar in meaning. We remark that for word embeddings there is no underlying data structure connecting different terms: instead, reasoning task can exploit the pairwise similarity between embeddings.

\subsection{Injecting Semantics in Image Classification Tasks}
In recent times, some research work has again proposed to integrate image classification with auxiliary information regarding the semantic context of labels. The most common procedure is to exploit structured representations of knowledge, such as ontologies, and extract from them a hierarchy of labels.
The existing approaches differ on how they incorporate class hierarchies in the training process: the literature distinguishes them among label embedding, hierarchical loss, and
hierarchical architecture-based methods.

Label-embedding approaches encode hierarchies directly into the class label representation using soft embedding vectors rather than one-hot encoding.
Some works directly use the taxonomic hierarchy tree of class to derive soft labels, also known as hierarchical embedding~\cite{barz2019hierarchy,mbm,bengio2010label,akata2015evaluation,xian2016latent,frome2013devise,liu2020hyperbolic}.
Barz and Denzler~\cite{barz2019hierarchy} derive a measure of semantic similarity between classes using the lowest common ancestor (LCA) height in a given hierarchy tree. The loss function combines two terms: a standard cross-entropy loss for the image classification target and a linear loss to enforce similarity between the image representations and the class semantic embedding.
Bertinetto et al.~\cite{mbm} also adopt the LCA to derive soft labels which encode the semantic information and use standard cross-entropy loss.
Using taxonomic hierarchy trees to encode hierarchies allows the direct use of prior semantic knowledge without needing external models or training procedures. On the other hand, it limits its application to problems for which a taxonomic hierarchy tree is available.
Other works proposed in the context of zero-shot classification avoid the use of taxonomies by exploiting text embeddings~\cite{akata2015evaluation,frome2013devise,xian2016latent}. For example, DeViSE~\cite{frome2013devise} derives class embedding using a pre-trained word2vec model on Wikipedia; these works were originally proposed in the context of zero-shot classification.
%
Few works explore both hierarchical and text embeddings. For example, Liu et al.~\cite{liu2020hyperbolic} combine the hyperbolic embedding learned with WordNet hierarchy with the word ones using the Glove model but for the context of zero-shot recognition.

Hierarchical loss methods integrate hierarchical semantic information by modifying the training loss function~\cite{mbm,deng2010does,haf,verma2012learning,wu2016learning} belong to this category.
Bertinetto et al.~\cite{mbm} propose a hierarchical cross-entropy (HXE) which is the weighted sum of the cross-entropies of the conditional probabilities derived by conditioning the predictions for a particular class on the parent-class probabilities.
Garg et al., in Hierarchy Aware Features (HAF)~\cite{haf}, propose a multi—term loss function to optimize fine label prediction while capturing hierarchical information. The approach leverages multiple classifiers, one at each level of the taxonomic tree hierarchy, with a shared feature space. In our work, we include semantic information directly via soft labels and also generalize for non-hierarchical semantic information.

Hierarchical architecture-based methods integrate the class hierarchy directly into the model architecture at the structural level. 
Approaches as ~\cite{chang2021your,redmon2017yolo9000,morin2005hierarchical} fall into this category. Hierarchical architecture-based methods are suitable for only hierarchy-based semantic information. We propose to encode class relations from generic semantic information, from hierarchy trees to word embeddings.

Our work proposes a label-embedding strategy that encodes semantics into class similarity embedding from general semantic information, be it taxonomic tree hierarchies or word embeddings. Moreover, differently than existing approaches, we explicitly address the assessment of the semantic-aware representation in the derived models.

\subsection{eXplainable and Interpretable Artificial Intelligence}
\label{subsec:xai}
Explainable Artificial Intelligence (XAI) refers to the ability of an AI system to explain its decisions and reasoning in a way that humans can easily understand. 
The need for XAI arises from the fact that many AI systems, particularly deep learning models, operate as black boxes, making it difficult for humans to understand how they arrive at their decisions~\cite{xai_survey}. 
There are two main approaches to achieving XAI: ex-post explainability and the design of intrinsically interpretable architectures. Ex-post explainability involves analyzing the outputs of a trained AI model and deriving explanations from them (~\cite{pastor-2019-explaining,lime2016,inputxgrad,saliency,integrated_gradients}, inter alia). This approach is commonly used with standard black-box models that lack inherent interpretability, such as deep neural networks. In contrast, interpretable architectures aim to be intrinsically (more) transparent from the outset. These models typically use simpler algorithms (such as decision trees or rule-based systems) as building blocks, or provide prototype-based explanations~\cite{rudin}.

\begin{figure}[h]
\includegraphics[width=\textwidth]{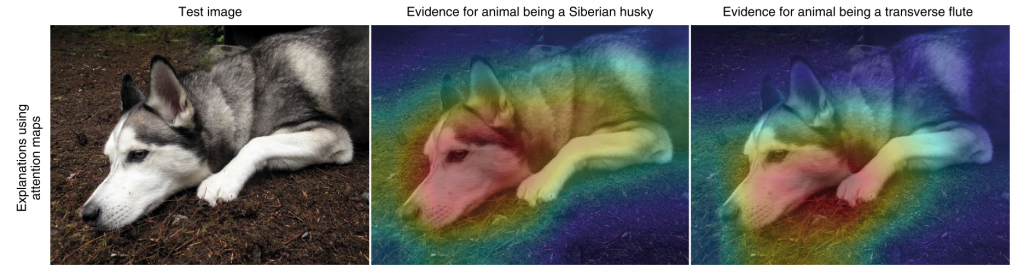}
\caption{Same model, same image, different classes, similar heatmaps~\cite{rudin}.}
\label{fig:huskyflute}
\end{figure}

In the specific case of explainability for image classification, it is worth mentioning that one of the driving domains is medical diagnosis~\cite{xai_medical}. From the algorithmic standpoint, most approaches belong to the ex-post category, and therefore extract explanation from already-trained, standard image classifiers.
One popular technique is \textit{Saliency Mapping~}\cite{saliency}, a method for computing the spatial support of a given class in a given image (image-specific class saliency map) using a single back-propagation pass through a classification convolutional model.
Other approaches are gradient-based attributions, such as \textit{Integrated Gradients}~\cite{integrated_gradients} and \textit{Input X Gradient}~\cite{inputxgrad}: these algorithms assign an importance score to each input feature by approximating the integral of gradients of the model’s output with respect to the inputs along the path (straight line) from given baselines / references to inputs. 
The common outputs of these systems are heatmaps - graphical representations of the importance of different regions in an image for the classification decision made by the neural network.
Heatmaps are typically overlayed on the original image, in order to better inspect the highlighted areas. 
Explaining image classifiers through heatmaps has notably received criticism~\cite{rudin}, as heatmaps explaining different classes often correspond to very similar heatmaps. For instance, in the famous example reported in Figure~\ref{fig:huskyflute}, Rudin et al.~\cite{rudin} showed how the heatmaps for \textit{Siberian Husky} and \textit{Transverse Flute} were very similar.

\section{Semantically-Augmented Labels}
\label{sec:semloss}
In this Section, we formally define our approach S-AL. We combine OHE ground truth vectors with auxiliary information, thus creating semantically augmented labels. Such labels can then be used in supervised image classification tasks in place of standard OHE vectors, so that no custom training loss or model architecture is necessary.

\subsection{Semantic-augmented Labels}

Consider an image classification dataset, with $X$ images and $Y$ labels, and assume that $\bar{y}$ indicates the OHE ground-truth label spanning over $C$ different classes (e.g. 10 in MNIST~\cite{mnist}, 1000 in ImageNet~\cite{imagenet}). Suppose we want to train a ML model in order to learn a mapping from $X$ to $Y$. Then, if for an image $x \in X$ the model predicts the vector $\hat{y}$, the loss function is $\textrm{Cross-Entropy} \, ( \bar{y}, \hat{y} ) = - \sum_{i \in C} \bar{y}(i) \odot \log(\hat{y}(i))$.


Now, let $e_j \in \mathbb{R}^{D}$ be a D-dimensional real vector containing auxiliary information for class $j \in C$. Let $EM \in \mathbb{R}^{C \times D}$ be the embedding matrix stacking $e_j, \ \forall j \in C$. We then compute the Gram matrix of the set of vectors $e_j$, i.e. $EM\cdot EM^T$, and call it the \textit{auxiliary information matrix}; furthermore, $\forall j \in C,$ we call $EM\cdot EM^T[j] = \tilde{y}$ the auxiliary label for class $j$. 
Every $\tilde{y}$ vector has $C$ elements and represents the similarities of the auxiliary information of the given class with all other auxiliary vectors.

In a first formulation, one can then define an enriched loss function as 
\begin{equation}\label{eq:1}
- \left[
\beta \, \sum_{i \in C} \bar{y}(i) \odot \log(\hat{y}(i)) +
(1 - \beta) \, \sum_{i \in C} \tilde{y}(i) \odot \log(\hat{y}(i))
\right]
\end{equation}
where the parameter $\beta$ governs the balance between standard cross-entropy (leftmost addendum) and the novel regularisation term pertaining to the auxiliary information (rightmost addendum).

However, the OHE labels $\bar{y}$ and the auxiliary labels $\tilde{y}$ can be combined into augmented labels $\yplus$ as follows:
\begin{equation}\label{eq:2}
\yplus \equiv \beta \, \bar{y} + (1 - \beta) \, \tilde{y}
\end{equation}
We can then merge Equation~\ref{eq:1} with Equation~\ref{eq:2}, thus producing the loss function:
\begin{equation}\label{eq:3}
-\sum_{i \in C} \yplus(i) \odot \log(\hat{y}(i)) 
\end{equation}
which corresponds to  Cross-Entropy$(\yplus,\hat{y})$. Therefore, we can enrich OHE ground truth vectors $\bar{y}$ with custom auxiliary information vectors $\tilde{y}$, producing semantically-augmented label vectors $\yplus$ that can be plugged as ground truth vectors for in a standard cross-entropy loss function, and exploit them, for instance, for an image classification training/learning process.


\begin{algorithm}
\caption{Generation of Semantically-Augmented Labels}\label{alg:salt}
\begin{algorithmic}[1]
\Require class number $C$, auxiliary vectors $e_1,...,e_C$
\State OHE matrix $\gets$ identity matrix $C \times C$
\State Embedding matrix EM $\gets$ Stack([$e_1,...,e_C$])
\State Auxiliary matrix AM $\gets EM\cdot EM^T$
\State Augmented matrix S-AL $\gets \beta$OHE + $(1-\beta)$AM
\State \Return S-AL
\end{algorithmic}
\end{algorithm}

Our approach is summarised in Algorithm~\ref{alg:salt}.
In a nutshell, we can collect auxiliary information about the labels, stack them in an embedding matrix $EM$ (line 2), then compute the Gram matrix of $EM$ (line 3) and use its vectors as auxiliary labels for each class. We then compute a weighted sum of OHE labels with the auxiliary matrix $AM$ and obtain semantically-augmented labels $\yplus \in \mathbf{R}^{C \times C}$ (line 4) that can be used downstream as ground truth in standard cross-entropy loss. We stress that our approach is general, as it exploits the pairwise similarities of auxiliary labels, regardless of whether they belong to an ontology or not. In order to exemplify our approach and drive the experimental section, in the next Subsections we will discuss how to generate S-AL for two selected sources of auxiliary information: label hierarchies and word embeddings.

\subsection{Exploiting Taxonomies}
\label{subsec:htal}
\begin{figure}[h]
\centering
\includegraphics[width=\textwidth]{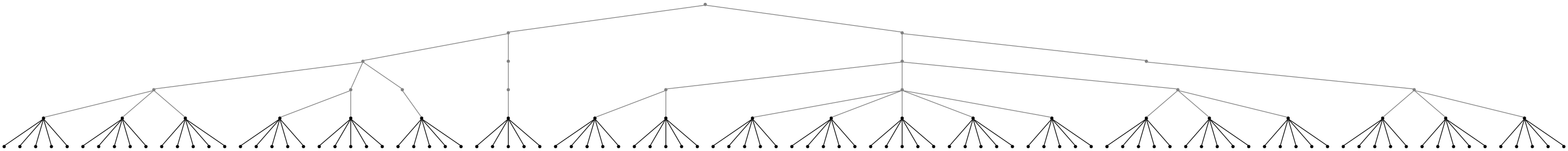}
\caption{Supporting taxonomy for the CIFAR100 labels.}
\label{fig:taxonomy}
\end{figure}

An ontology is a description of classes, properties, and relationships in a domain of knowledge. If the labels of the image classification task can be organised in an ontology, their relative position in the data structure can provide additional information about the relative similarity of the semantic concepts  - e.g., one would expect the path from \textit{spoon} to \textit{tea-spoon} to be considerably shorter than the path from \textit{spoon} to \textit{shark}. A classical example of this link from the neural domain to the symbolic one is the aforementioned ImageNet-Wordnet link: every ImageNet label is a node in the Wordnet ontology. In principle, this allows to exploit semantic information when training or explaining ImageNet-based image classification ML models; however, this resourceful connection has seldom been exploited. In this paper, we focus on  CIFAR100\unskip\footnote{http://www.cs.toronto.edu/$\sim$kriz/cifar.html}, a common image classification benchmarking dataset with the additional feature that its labels are connected in a taxonomy, outlined in Figure~\ref{fig:taxonomy}.

CIFAR100 includes 100 classes, represented by the leaves of the tree. The authors of the dataset further group the 100 labels into 20 five-sized macro-classes, also called \textit{coarse labels} or \textit{superclasses}. For instance, the labels \textit{maple, oak, palm, pine, willow} are clustered together in the macro-class \textit{trees}. This original two-layered taxonomy corresponds to the two lowest layers of Figure~\ref{fig:taxonomy}. The taxonomy was further extended with additional layers, corresponding to even larger groupings of labels. The resulting taxonomy encompasses 100 level-0 labels, 20 level-1 labels, 8 level-2 and 4 level-3 labels, and 2 level-4 labels, with the single level-5 label corresponding to the \textit{root} node of the taxonomy. Throughout the paper, we will often refer to hierarchical \textit{depths} or \textit{levels}.

\begin{figure}[h]
\includegraphics[width=\textwidth]{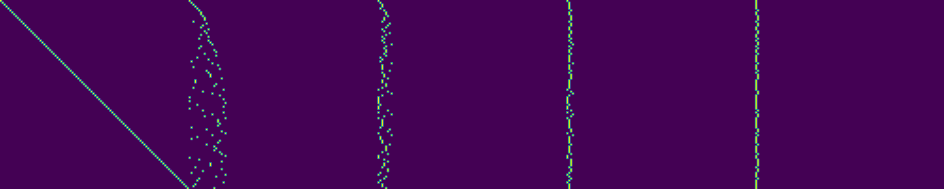}
\caption{Taxonomy-derived embedding matrix $EM$ for the CIFAR100 labels.}
\label{fig:hier_gt}
\end{figure}

In order to extract auxiliary information vectors from this taxonomy, we first extract the leave-to-root path of each class, and then stack the OHE of every step in the path. For instance, a class with path 23-45-5 will be encoded as the concatenation of three OHE vectors, with hot elements in position 23, 45 and 5 respectively. For CIFAR100, the resulting embedding matrix $EM$ is depicted in Figure~\ref{fig:hier_gt}.

This matrix has 100 rows, encoding all classes in $C$. Since the hierarchy has 5 levels, we obtain an embedding dimension $D$ of 500, due to the stacking of 5 100-units OHE vectors. One can observe that the leftmost part of $EM$ corresponds to the identity matrix of the single classes, whereas the rightmost part has only hot elements in the first two columns, since the level-4 labels can only be 0 or 1. As a minor implementation detail, we remark that the zero-padding will have no influence when computing the Gram matrix of $EM$ in order to obtain the semantically-augmented labels. 

\begin{figure}[h]
\begin{subfigure}[b]{.49\linewidth}
\centering
\includegraphics[width=.8\textwidth]{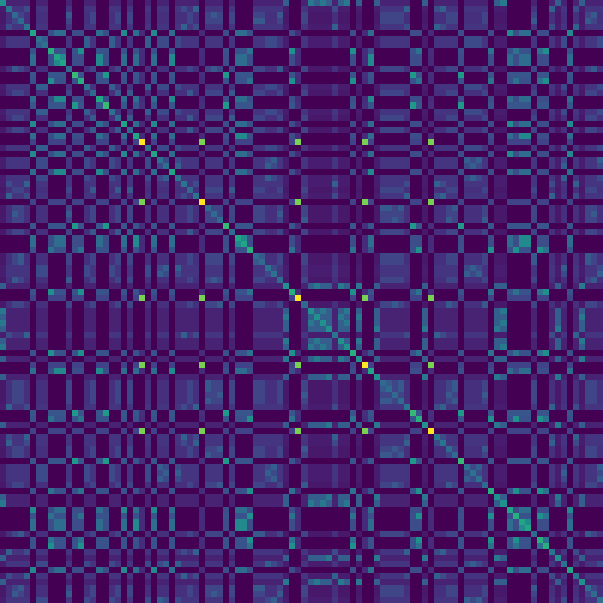} 
\caption{}\label{fig:hier_gram:a}
\end{subfigure}
\begin{subfigure}[b]{.49\linewidth}
\centering
\includegraphics[width=.8\textwidth]{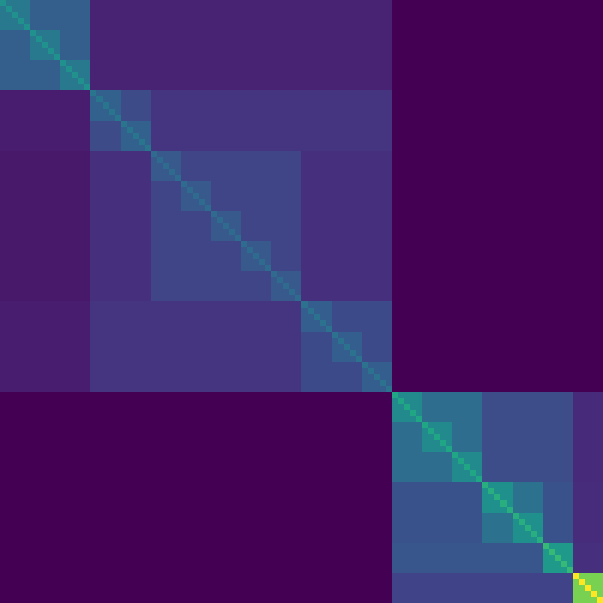} 
\caption{}\label{fig:hier_gram:b}
\end{subfigure}
\caption{Taxonomy-derived S-AL matrix for the CIFAR100 labels. The original class order (a) can be taxonomically sorted (b) for visual inspection.}
\label{fig:hier_gram}
\end{figure}

For CIFAR100, the auxiliary labels $EM\cdot EM^T$ are computed from the embedding matrix $EM$ depicted in Figure~\ref{fig:hier_gt}. By combining the auxiliary labels with the OHE labels, the resulting S-AL matrix is depicted in Figure~\ref{fig:hier_gram:a}.

Re-arranging the CIFAR100 classes for semantic similarity (Figure~\ref{fig:hier_gram:b}), one can visually inspect the S-AL. We observe how these labels capture similarities at various hierarchical depths, represented by coloured blocks of different size. The diagonal represent self-similarities, while the small regular five-by-five blocks along the diagonal represent the similarities within macro-groups of labels (corresponding to the coarse labels in CIFAR100, such as the \textit{tree} case mentioned above). Larger blocks correspond to higher levels in the taxonomy. The S-AL depicted in Figure~\ref{fig:hier_gram:a} will be used as ground truth to train image classifiers in the experimental part of the paper. We remark that the re-arrangement is for visualisation purposes, and that the matching between class index and S-AL is never modified.

\begin{figure}[h]
\resizebox{\textwidth}{!}{
\centering \input{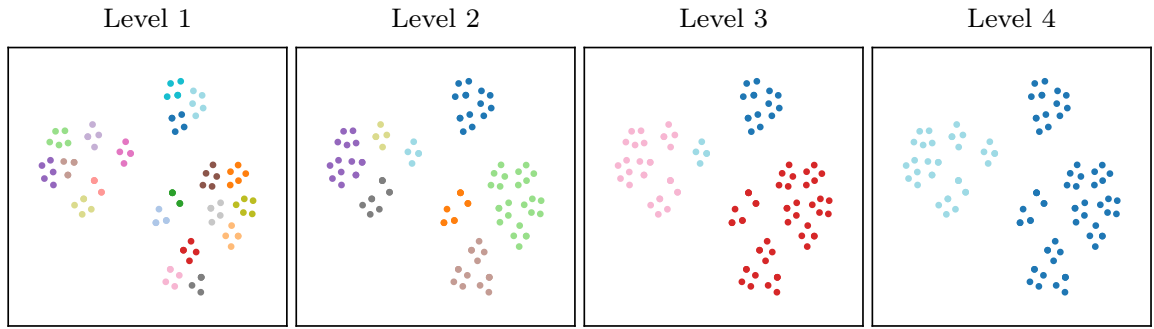}}
\caption{t-SNE compression of the S-AL for CIFAR100 classes.}
\label{fig:hier_tsne}
\end{figure}

For further visual inspection, in Figure~\ref{fig:hier_tsne} we report a t-SNE~\cite{tsne} embedding of the hierarchy-augmented labels for CIFAR100. t-SNE (\textit{t-distributed Stochastic Neighbour Embedding}) is an unsupervised dimensionality reduction technique for embedding high-dimensional data in a low-dimensional space of two dimensions. 

The color-coding is applied afterwards and corresponds to labels at different hierarchical depths, going from level-1 (original 100 CIFAR100 coarse labels) to level-4 (two macro-groups below the root in Figure~\ref{fig:taxonomy}). At level-0 all data points would be coloured with 100 independent colours, whereas level-5 would be monochromatic, so we omit these two panels. Instead, we depict the color-coding for all other depths in the taxonomy, showing how S-AL display similarities at every aggregation level. The intuition that will drive our experiments is that these augmented labels can be a better ground truth, with respect to simple/standard OHE vectors, for image classification tasks.

\subsection{Exploiting Word Embeddings}
\label{subsec:weal}
It is not always the case that labels belong to an ontology or can be arranged in a hierarchical taxonomy. Our approach allows to exploit any auxiliary information about a label, provided that it can be expressed as a vector. We argue that Word Embeddings represent the perfect candidate for this: it is straightforward to obtain vector representations for labels, and the Euclidean distance (or cosine similarity) between two word vectors provides an effective method for measuring the linguistic or semantic similarity of the corresponding words. 

\begin{figure}[h]
\includegraphics[width=\textwidth]{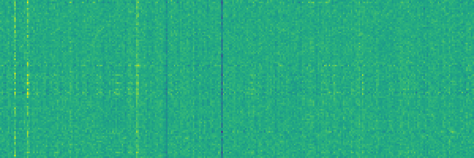}
\caption{GloVe-derived embedding matrix $EM$ for the CIFAR100 labels.}
\label{fig:glove_gt}
\end{figure}

\begin{figure}[h]
\begin{subfigure}[b]{.49\linewidth}
\centering
\includegraphics[width=.78\textwidth]{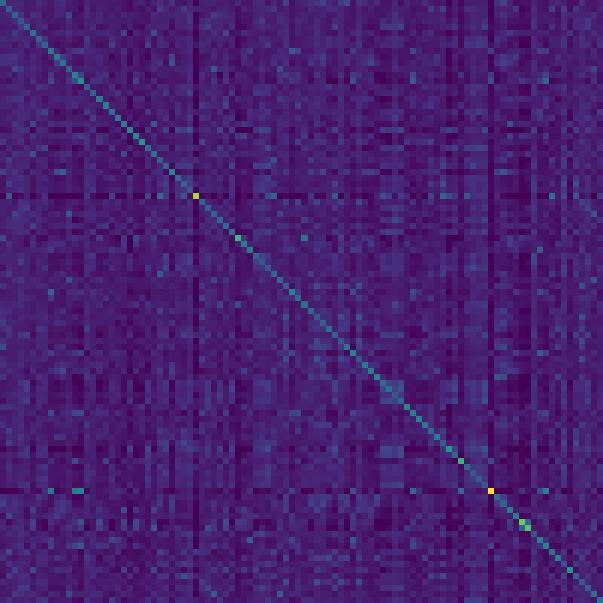}
\caption{}\label{fig:glove_gram:a}
\end{subfigure}
\begin{subfigure}[b]{.49\linewidth}
\centering
\includegraphics[width=.78\textwidth]{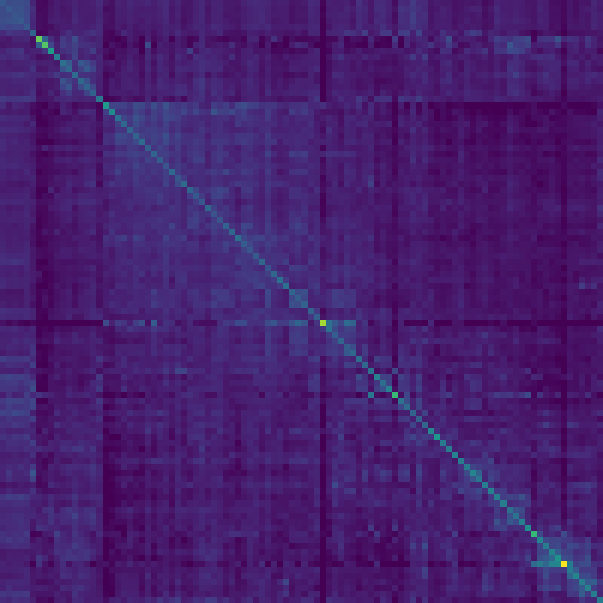}
\caption{}\label{fig:glove_gram:b}
\end{subfigure}
\caption{GloVe-derived S-AL matrix for the CIFAR100 labels. }
\label{fig:glove_gram}
\end{figure}

In this paper we exploit GloVe~\cite{glove} (\textit{Global Vectors for Word Representation}) to obtain the vector representations of all 100 classes of CIFAR100. The resulting augmented labels are depicted in Figure~\ref{fig:glove_gt}. One can observe how these vectors are denser, compared to the ones extracted from the ontology and displayed in Figure~\ref{fig:hier_gt}. However, we follow the same procedure outlined in Algorithm~\ref{alg:salt}, thus obtaining a S-AL matrix that we will later use as image classification ground truth for the experimental phase. We remark that in this case we are not tapping into CIFAR00's provided taxonomy - instead, we are interested in exploring the contextual relationship between labels (emerging from GloVe embeddings) when paired with the visual similarities that occurs in CIFAR100 images. 
The GloVe-derived S-AL for CIFAR100 are depicted in Figure~\ref{fig:glove_gram:a}. In Figure~\ref{fig:glove_gram:b}, we show a different class arrangement, defined through a hierarchical clustering procedure. We remark again that this is merely for visualisation purposes, and has no impact on the downstream training task.

As for the taxonomical case, we ran our S-AL through t-SNE in order to visually inspect whether they preserve some structure at different hierarchical depths. We remark that, in this case, the hierarchical structure is not part of the augmented labels, which are computed from OHE labels and GloVe embeddings; instead, we use hierarchical levels to visualise and inspect the obtained S-AL. 

\begin{figure}[h]
\resizebox{\textwidth}{!}{
\centering \input{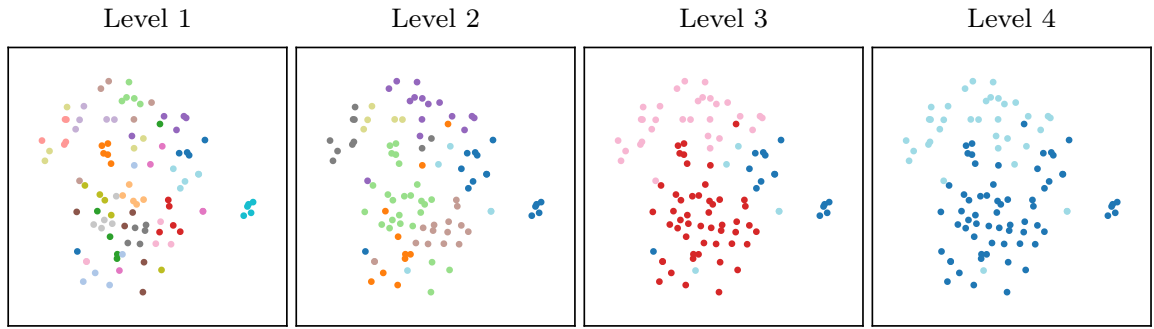}}
\caption{t-SNE visualisation of the GloVe-augmented labels}
\label{fig:glove_tsne}
\end{figure}

It is not surprising that the t-SNE embedding of GloVe augmented labels in Figure~\ref{fig:glove_tsne} is less clustered when compared to the hierarchical labels in Figure~\ref{fig:hier_tsne}, but at the same time one can appreciate the structure emerging at levels 3 and 4. These augmented labels derive from word embeddings, so their relative similarities might follow a different logic with respect to the hierarchical ones: for instance, in panel 3 there is a lonesome light blue dot in the lower half, surrounded by red ones. That dot is \textit{sea}, and amongst the surrounding red dots we have \textit{aquarium\_fish, crab, crocodile, dolphin, flatfish, otter, shark, trout, turtle, whale}. These marine (or at least aquatic) animals are taxonomically very different from \textit{sea}, and this is captured by the different colour-coding, but they clearly share semantic similarities with \textit{sea} and the proximity of these points in the (t-SNE reduced) S-AL space shows how the semantic information extracted from GloVe embeddings carries this information.

\FloatBarrier
\section{Experimental Settings}
\label{sec:exp_setup}

In this section we describe the experimental settings that we adopted to train models and evaluate the impact of injecting knowledge in the image classification process through S-AL. 

\subsection{Data and Models}
As mentioned in the previous Section, we focus on CIFAR100, an image classification dataset commonly used as benchmark. We exploited the CIFAR100 taxonomy and GloVe embeddings of the labels to generate S-AL; furthermore, we refer to the taxonomy in Figure~\ref{fig:taxonomy} when discussing depths/levels.

The set of image classifiers we trained for our experimental phase is the following:
\begin{itemize}
    \item \textbf{XENT}: we trained a Wide ResNet model with standard cross-entropy, to be used as a reference point for fine-grained image classification accuracy, and as a baseline for all other evaluations.
    \item \textbf{HAF} 
    refers to the Hierarchy Aware Features (HAF) model as described in Garg et al.~\cite{haf}. We used the available code and re-trained the model for comparison with S-AL when dealing with taxonomically-enriched labels. However, we remark that HAF is explicitly trained on the hierarchy, and therefore cannot be used for comparison in the GloVe-based set of experiments.
    \item \textbf{SOFT*} refers to the soft-label approach of Bertinetto et al.~ \cite{mbm}. 
    As we did for HAF, we adopted already used values for the hyperparameters, in order to reproduce the original results. For SOFT*, we trained two parametrised versions, SOFT4 and SOFT30. 
    Also SOFT* is limited to be used for the taxonomical case only.
    \item \textbf{HT-AL*} is a family of models trained with our S-AL approach, where the auxiliary information is obtained from the hierarchical taxonomy (HT), as discussed in Subsection~\ref{subsec:htal}. For every model HT-AL\textit{x}, \textit{x} indicates the value of $\beta$ used to compute the S-AL.
    \item \textbf{WE-AL*}, analogously, is a family of models trained with our S-AL approach, where the auxiliary information is obtained from a word embedding (WE), as discussed in Subsection~\ref{subsec:weal}.
\end{itemize}

The synopsis of these models is summarised in Table~\ref{tab:models}.

\begin{table}[h]
\begin{center}
\begin{tabular}{ |c|c|c| } 
 \hline
\textbf{Model(s)} & \textbf{Label} & \textbf{Notes}  \\ 
 \hline
 \hline
XENT & OHE & Baseline  \\ 
 \hline
SOFT4 & Hierarchy & Cannot encode embeddings \\ 
 \hline
 SOFT30 & Hierarchy & Cannot encode embeddings \\ 
 \hline
HAF & Hierarchy & Cannot encode embeddings \\ 
 \hline
HT-AL$\beta$ & Hierarchy-augmented & Our approach   \\ 
 \hline
WE-AL$\beta$ & GloVe-augmented & Our approach   \\ 
 \hline
\end{tabular}
\vspace*{2mm}
\caption{Overview of models of the experimental phase}
\label{tab:models}
\end{center}
\end{table}

We briefly report how, while we kept all architecture models as similar as possible, including the number of training epochs, we observed how HAF required more training time with respect to all other models (with a factor spanning from x3 to x4). We conjecture that this is due to
the higher complexity of the architecture (multiple classification heads) and the fact that the training loss is composed by four independent terms.



\subsection{Image Classification}
The very first metric to be taken into account is, obviously, classification accuracy - that is, the percentage of (out-of-sample) datapoints that a model is able to correctly classify. We extend this metric by evaluating it at every ontological depth: for instance, since {\em maple} and {\em oak} (level-0 labels) are both {\em trees} (level-1 label), a {\em maple} image which is classified as {\em oak} will be considered a level-0 error, but a level-1 correct classification. 

Second, we measure the average mistake severity, a performance measure introduced by Bertinetto et al.~\cite{mbm}. The mistake severity takes into account misclassified images, and measures the \textit{lowest common ancestor} (LCA) distance between true and predicted labels; the average LCA across all misclassifications is then reported. This error metric ignores the \textit{quantity} of mistakes made by the model, trying  to characterise their \textit{quality} instead. Clearly, this entails that if model A classifies all images correctly but one, and in that single case the semantic distance is high (e.g. \textit{maple} - \textit{rocket}), and model B misclassifies all images but always predicting ontologically similar classes (e.g. \textit{maple} - \textit{oak}), the mistake severity will rank B as the best method of the two. Therefore, we argue that mistake severity, albeit informative, should also be paired with other metrics, such as explainability.

\subsection{Representation Learning}
We are interested in investigating whether S-AL impacts the inner knowledge representation learned by a ML model. To do so, we use all models as feature extractors and analyse the feature space on which the out-of-sample images are projected. Intuitively, standard OHE-supervised training forces the ML model to project all images of one class to a compact point cloud, but there is nothing enforcing that point clouds of similar classes should be close to each other. Our hope is that, by injecting semantic information, the compactness of single point clouds is preserved, but the feature space is reorganised in such a way that semantically similar classes are geometrically close. To measure this, for each taxonomy level $x$ we assign each feature vector to its corresponding level-$x$ label - e.g., for level-1 we label \textit{trees} each feature vector corresponding to images of maples, oaks, etc.. We then run several clustering evaluation metrics on the resulting partition system to inspect the compactness of the emerged clusters.


\subsection{Explainability}
Finally, we test the impact of our approach on the produced explanations. We conduct our experiments in order to inspect whether S-AL, by injecting auxiliary information in the image classification process, manages to mitigate the husky-flute effect mentioned in Section~\ref{subsec:xai}.
Ideally, the similarity of two heatmaps should be proportional to the ontological proximity of the two classes they explain.

We exploiting three standard explainers (\textit{Integrated Gradients}, \textit{Input-X Gradient}, and \textit{Saliency}), all available within the Captum library\unskip\footnote{https://captum.ai/}. We then produce explanations (heatmaps) for every image in the CIFAR100 test set, for every ML model listed above, and every output neuron, corresponding to a class. We are interested in comparing the true class, and true class heatmap, with all other classes. For instance, referring again to the famous example in Figure~\ref{fig:huskyflute}, we are interested in comparing the classes of \textit{husky} and \textit{flute}, and the heatmaps produced for \textit{husky} and for \textit{flute}. For brevity, we will refer to the true class heatmap as \textit{true\_heatmap} and to any currently explained class heatmap as \textit{expl\_heatmap}.

We rely on the CIFAR100 ontology to define the semantic pairwise distance between classes. We observe that there is no consensus regarding metrics for heatmap comparison or benchmarking, and we therefore implemented several custom functions:
\begin{itemize}
    \item \textbf{Mean Absolute Difference}: average absolute value of the per-pixel difference between \\ \textit{true\_heatmap} and \textit{expl\_heatmap}.
    \item \textbf{Deletion Curve Distance}: using the \textit{true\_heatmap} to rank pixels, spanning from the highest-scoring to the lowest-scoring locations, we progressively remove elements from both \textit{true\_heatmap} and \textit{expl\_heatmap}. At each step we compute the sum of the two remaining heatmaps, and compare the two resulting curves. This metric is inspired by the Deletion Curve~\cite{rise}.
    \item \textbf{Spearman Distance}: we compute the Spearman-ranking correlation between the two heatmaps, normalising it in the 0-1 range.
    \item \textbf{Progressive Binarisation}: we extract a set of progressive thresholds from the \textit{true\_heatmap} and use them to binarise both heatmaps. We then check the intersection of the resulting pairs of masks at each step.
\end{itemize}





\section{Experimental Results}
\label{sec:exp_res}
\subsection{Image Classification}
\begin{table}
\begin{tabularx}{\textwidth}{llXXXX}
\toprule
 &  & XENT & HAF & HT-AL & WE-AL \\
Level &  &  &  &  &  \\
\midrule
\multirow[t]{3}{*}{0} & Error@1 & 22.462 ± 0.283 & 22.420 ± 0.203 & 22.718 ± 0.267 & 22.206 ± 0.430 \\
 & Error@5 & 6.234 ± 0.118 & 6.414 ± 0.230 & 7.398 ± 0.148 & 6.314 ± 0.080 \\
 & MS & 2.364 ± 0.025 & 2.251 ± 0.018 & 2.218 ± 0.021 & 2.307 ± 0.022 \\
\cline{1-6}
\multirow[t]{3}{*}{1} & Error@1 & 14.002 ± 0.170 & 13.470 ± 0.188 & 13.554 ± 0.153 & 13.598 ± 0.396 \\
 & Error@5 & 3.058 ± 0.086 & 3.966 ± 0.187 & 5.218 ± 0.114 & 3.556 ± 0.151 \\
 & MS & 2.188 ± 0.027 & 2.082 ± 0.029 & 2.042 ± 0.023 & 2.134 ± 0.027 \\
\cline{1-6}
\multirow[t]{3}{*}{2} & Error@1 & 9.182 ± 0.184 & 8.372 ± 0.107 & 8.312 ± 0.128 & 8.652 ± 0.283 \\
 & Error@5 & 1.322 ± 0.032 & 2.256 ± 0.143 & 3.672 ± 0.064 & 1.770 ± 0.154 \\
 & MS & 1.812 ± 0.026 & 1.741 ± 0.030 & 1.698 ± 0.020 & 1.782 ± 0.013 \\
\cline{1-6}
\multirow[t]{3}{*}{3} & Error@1 & 4.488 ± 0.108 & 3.780 ± 0.129 & 3.552 ± 0.119 & 4.088 ± 0.167 \\
 & Error@5 & 0.396 ± 0.030 & 0.976 ± 0.093 & 2.238 ± 0.093 & 0.790 ± 0.095 \\
 & MS & 1.661 ± 0.020 & 1.641 ± 0.034 & 1.635 ± 0.018 & 1.656 ± 0.012 \\
\cline{1-6}
\multirow[t]{3}{*}{4} & Error@1 & 2.966 ± 0.131 & 2.424 ± 0.177 & 2.254 ± 0.094 & 2.682 ± 0.144 \\
 & Error@5 & 0.196 ± 0.021 & 0.626 ± 0.087 & 1.660 ± 0.091 & 0.404 ± 0.066 \\
 & MS & 1.000 ± 0.000 & 1.000 ± 0.000 & 1.000 ± 0.000 & 1.000 ± 0.000 \\
\cline{1-6}
\bottomrule
\end{tabularx}
\vspace*{2mm}
\caption{Error and MS, at all hierarchical depths, for our selected models.}
\label{tab:ml_sel}
\end{table}
\begin{figure}[h!]
\resizebox{\textwidth}{!}{
\centering \input{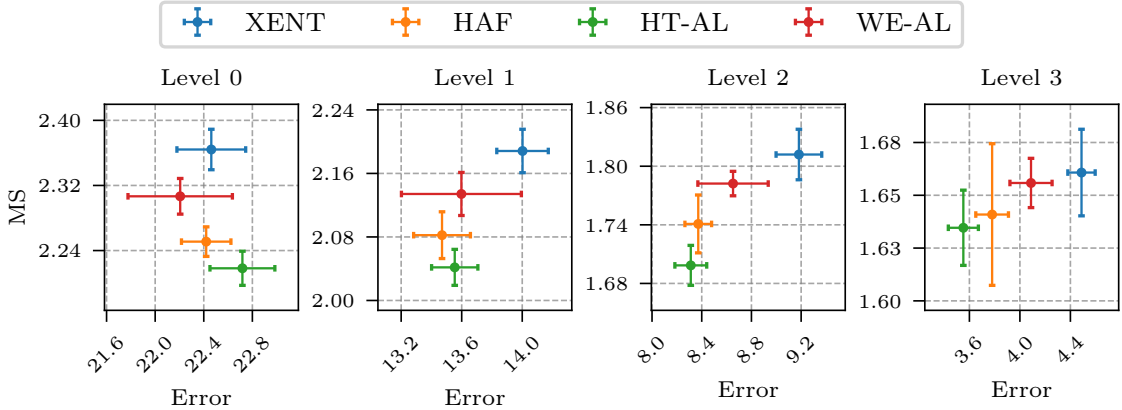}}
\caption{Error and MS, ad different hierarchical depths, for the selected models.}
\label{fig:ml_sel}
\end{figure}
\begin{table}[h!]
\begin{tabularx}{\textwidth}{l|XXXXXX}
\toprule
 & Error@1 & Error@5 & MS & HD@1 & HD@5 & HD@20 \\
\midrule
XENT & 22.462 & 6.234 & 2.364 & 0.531 & 2.242 & 3.178 \\
SOFT-4 & 32.580 & 16.480 & 2.206 & 0.719 & 1.235 & 2.223 \\
SOFT-30 & 26.650 & 8.940 & 2.331 & 0.621 & 1.375 & 2.798 \\
HAF & 22.420 & 6.414 & 2.251 & 0.505 & 1.422 & 2.640 \\
HT-AL $(\beta = 0.30)$ & 23.390 & 7.690 & 2.237 & 0.523 & 1.233 & 2.195 \\
HT-AL $(\beta = 0.35)$ & 22.820 & 7.510 & 2.252 & 0.514 & 1.236 & 2.195 \\
HT-AL $(\beta = 0.40)$ & 22.718 & 7.398 & 2.218 & 0.504 & 1.233 & 2.192 \\
HT-AL $(\beta = 0.45)$ & 22.760 & 7.060 & 2.219 & 0.505 & 1.242 & 2.194 \\
HT-AL $(\beta = 0.50)$ & 22.710 & 7.240 & 2.262 & 0.514 & 1.256 & 2.199 \\
WE-AL $(\beta = 0.30)$ & 23.230 & 7.640 & 2.303 & 0.535 & 1.836 & 2.808 \\
WE-AL $(\beta = 0.35)$ & 23.270 & 7.240 & 2.312 & 0.538 & 1.841 & 2.817 \\
WE-AL $(\beta = 0.40)$ & 23.260 & 7.040 & 2.311 & 0.538 & 1.851 & 2.822 \\
WE-AL $(\beta = 0.45)$ & 22.870 & 6.920 & 2.320 & 0.530 & 1.849 & 2.815 \\
WE-AL $(\beta = 0.50)$ & 22.760 & 6.820 & 2.291 & 0.521 & 1.864 & 2.823 \\
WE-AL $(\beta = 0.55)$ & 22.380 & 6.820 & 2.332 & 0.522 & 1.879 & 2.824 \\
WE-AL $(\beta = 0.60)$ & 22.490 & 6.490 & 2.349 & 0.528 & 1.887 & 2.833 \\
WE-AL $(\beta = 0.65)$ & 22.450 & 6.290 & 2.325 & 0.522 & 1.907 & 2.834 \\
WE-AL $(\beta = 0.70)$ & 22.206 & 6.314 & 2.307 & 0.512 & 1.914 & 2.843 \\
WE-AL $(\beta = 0.75)$     & 22.270 & 6.420 & 2.302 & 0.513 & 1.931 & 2.853\\ \bottomrule
\end{tabularx}
\vspace*{2mm}
\caption{Accuracy and mistake severity for all trained ML models.}
\label{tab:ml_all}
\end{table}
Our first goal is to verify how S-AL models perform in terms of image classification error quantity and quality: we report our results in Table~\ref{tab:ml_all}.
Each row indicates a model; the columns indicate the top-1 and top-5 classification error percentage (\textit{Error@1, Error@5}), the average hierarchical distance between true and predicted labels for misclassifications, also called \textit{mistake severity} (\textit{MS}), and the average hierarchical distance when taking into account the top-k predictions (\textit{HD@1, HD@5, HD@20}). We remark that $MS$ takes only into account misclassified data points, while $HD@k$ includes all cases. We thus successfully reproduced~\cite{mbm,haf} the results of XENT, HAF and SOFT*.
We observe that SOFT4 and SOFT30 do not produce competitive results in terms of accuracy, and we therefore exclude them for further experiments. HAF, on the other hand, yields competitive scores and will be kept. Regarding our models, reasonable choices for $\beta$ seem to be $0.4$ for HT-AL and $0.7$ for WE-AL: from now on we will therefore conduct experiments with HT-AL($\beta = 0.4$) and WE-AL($\beta = 0.7$), which we will simply indicate as HT-AL and WE-AL.

We therefore select XENT, HAF, HT-AL and WE-AL. For these models we run multiple experiments with different seeds in order to obtain error bars. We report the results in the first row of Table~\ref{tab:ml_sel}, corresponding to level-0. Furthermore, for the selected models we report error and MS at different hierarchical depths.

The results of Table~\ref{tab:ml_sel} can be also visualised as scatterplots, and we do so in Figure~\ref{fig:ml_sel}.

In accordance with the previous Sections, each panel represents a specific depth, or level, in the CIFAR100 hierarchy. We omit the panel of level-5, since it corresponds to the hierarchy root, and level-4, since at that level there are only two sub-groups, and therefore the mistake severity has constant value of 1. Concerning accuracy (\textit{error}, x-axis), we observe how all alternative models (HT-AL, WE-AL and HAF) perform comparably to XENT at level-0 (fine labels, standard classification), but outperform it for all other levels. Concerning MS, the model ranking is the same at all depths: HT-AL performs best, followed by HAF, then WE-AL, and finally XENT. We remark that HT-AL and HAF were trained on labels derived from CIFAR100, while WE-AL was not. We observe how HAF's MS error bar widens as the levels progress. In general, these experiments confirm our expectation that the injection of auxiliary knowledge can mitigate the hierarchical distance of misclassifications of a ML model at all hierarchical levels while maintaining competitive accuracy results.
\subsection{Representation Learning}
\begin{figure}[t!]
\centering
\includegraphics[width=.7\textwidth]{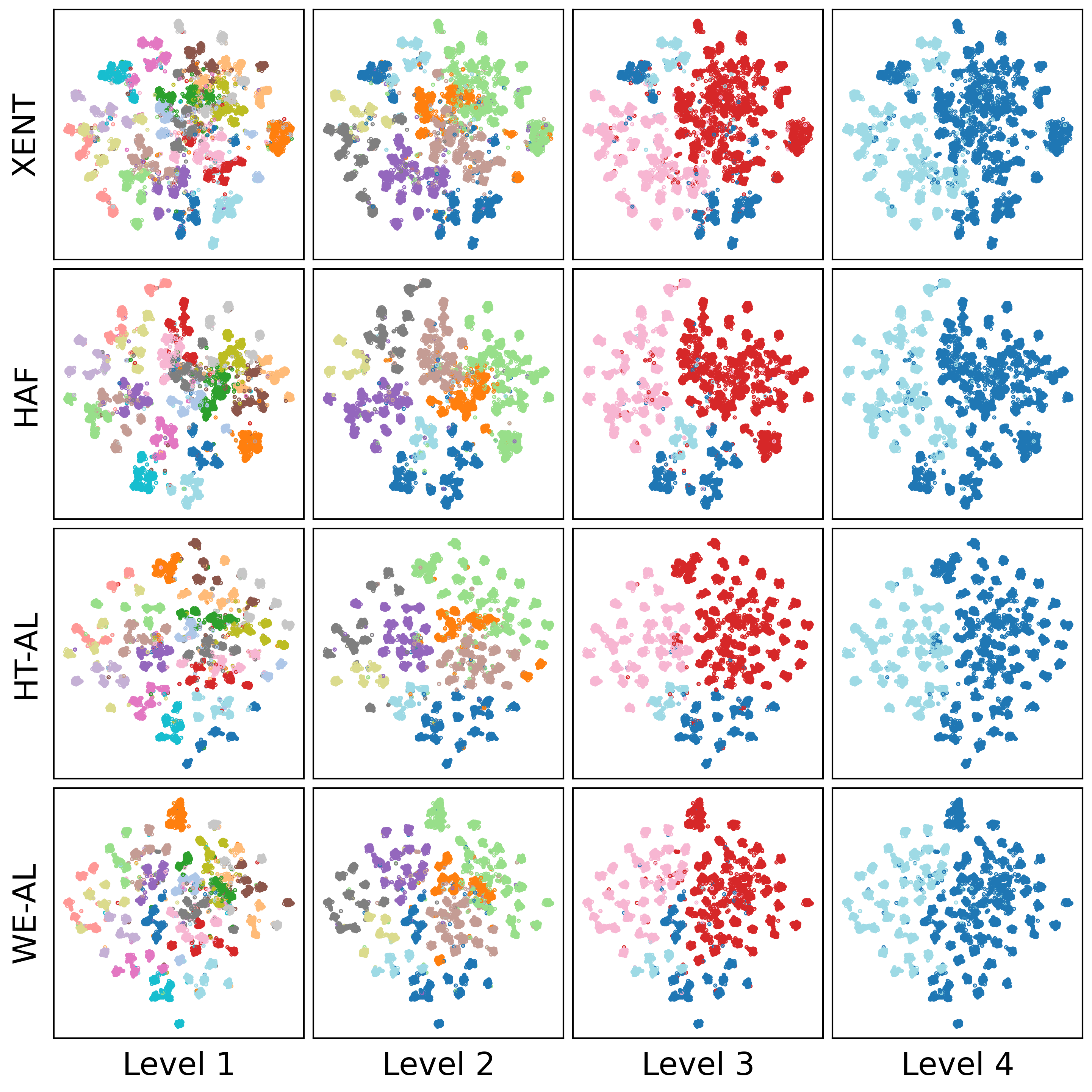}
\caption{t-SNE compression of the feature vectors extracted from the CIFAR100 test set. Colour-coded afterwards according to level-$k$ labels.}
\label{fig:feat_sne}
\end{figure}
Besides confirming that S-AL produces models which are competitive in terms of classification accuracy, we are also interested in peeking into the internal representation of knowledge that they have learned. In order to do so, we use our selected trained models XENT, HAF, HT-AL and WE-AL as feature extractors, mapping all images of the test set of CIFAR100 into a 512-dimensional latent feature space. As a first experiment, we use t-SNE again to reduce the dimensionality of the feature space, so that it can be visually inspected. We color-code all data points according to their label at different hierarchical levels, and report the resulting scatterplot in Figure~\ref{fig:feat_sne}. 

\begin{figure}[t!]
\resizebox{\textwidth}{!}{
\centering \input{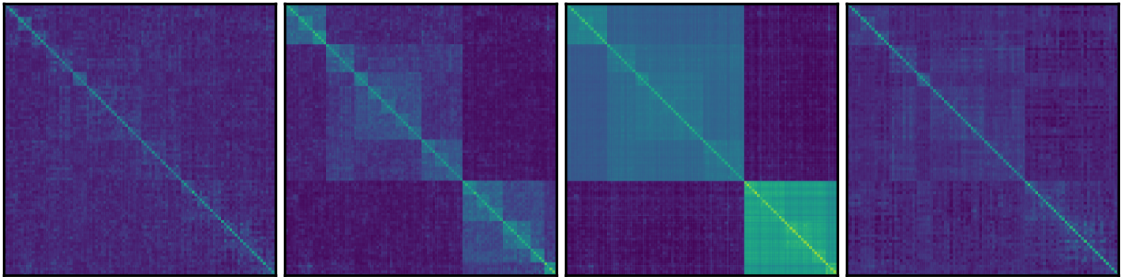}}
\caption{Visual inspection of the feature spaces for the selected models. From left to right: X\textbf{}ENT, HAF, HT-AL, and WE-AL.}
\label{fig:feat_dist}
\end{figure}

As a second step of analysis, we compute the pairwise cosine similarities between feature vectors, and report it in Figure~\ref{fig:feat_dist}. We observe how XENT shows no emerging structure besides the diagonal, HAF displays similarity blocks at different depths. Also HT-AL displays similarity blocks, but with strong visual importance to level-4 (two macro-blocks); finally, WE-AL shows shows a visually weaker structure.

\begin{table}[h!]
\begin{tabularx}{\textwidth}{ll|l|l|l|l}
\toprule
 &  & XENT & HAF & HT-AL & WE-AL \\
Level & Metric &  &  &  &  \\
\midrule
\multirow[t]{3}{*}{0} & Silhouette $(\uparrow)$ & 0.211 ± 0.002 & 0.232 ± 0.002 & 0.280 ± 0.003 & 0.299 ± 0.004 \\
 & Calinski-Harabasz $(\uparrow)$ & 162.5 ± 1.3 & 271.4 ± 3.1 & 298.3 ± 3.2 & 181.3 ± 2.2 \\
 & S-Dbw $(\downarrow)$ & 0.613 ± 0.001 & 0.512 ± 0.002 & 0.493 ± 0.002 & 0.588 ± 0.002 \\
\cline{1-6}
\multirow[t]{3}{*}{1} & Silhouette $(\uparrow)$ & 0.083 ± 0.001 & 0.163 ± 0.002 & 0.110 ± 0.000 & 0.100 ± 0.001 \\
 & Calinski-Harabasz $(\uparrow)$ & 178.1 ± 1.7 & 552.8 ± 4.4 & 583.9 ± 3.7 & 185.4 ± 0.9 \\
 & S-Dbw $(\downarrow)$ & 0.863 ± 0.001 & 0.698 ± 0.001 & 0.688 ± 0.001 & 0.859 ± 0.001 \\
\cline{1-6}
\multirow[t]{3}{*}{2} & Silhouette $(\uparrow)$ & 0.052 ± 0.001 & 0.156 ± 0.001 & 0.098 ± 0.001 & 0.059 ± 0.001 \\
 & Calinski-Harabasz $(\uparrow)$ & 221.0 ± 2.0 & 841.8 ± 3.9 & 1200.1 ± 7.4 & 250.9 ± 1.6 \\
 & S-Dbw $(\downarrow)$ & 0.918 ± 0.001 & 0.763 ± 0.001 & 0.717 ± 0.001 & 0.906 ± 0.001 \\
\cline{1-6}
\multirow[t]{3}{*}{3} & Silhouette $(\uparrow)$ & 0.043 ± 0.001 & 0.146 ± 0.001 & 0.121 ± 0.002 & 0.050 ± 0.002 \\
 & Calinski-Harabasz $(\uparrow)$ & 253.9 ± 4.2 & 1040.1 ± 5.4 & 2239.8 ± 11.3 & 327.6 ± 4.5 \\
 & S-Dbw $(\downarrow)$ & 0.933 ± 0.002 & 0.802 ± 0.002 & 0.736 ± 0.003 & 0.922 ± 0.001 \\
\cline{1-6}
\multirow[t]{3}{*}{4} & Silhouette $(\uparrow)$ & 0.037 ± 0.001 & 0.136 ± 0.002 & 0.311 ± 0.002 & 0.050 ± 0.001 \\
 & Calinski-Harabasz $(\uparrow)$ & 302.6 ± 8.9 & 1519.2 ± 19.1 & 5111.6 ± 38.5 & 467.4 ± 6.9 \\
 & S-Dbw $(\downarrow)$ & 0.989 ± 0.001 & 0.917 ± 0.000 & 0.799 ± 0.000 & 0.975 ± 0.001 \\
\cline{1-6}
\bottomrule
\end{tabularx}
\vspace*{2mm}
\caption{Clustering evaluation for all feature spaces.}
\label{tab:clustering}
\end{table}

However, besides intuitive visual inspections, we are interested in quantitatively characterising the feature space of each model. In order to do so, for each model and hierarchy level $k$, we partition the feature vectors into clusters according to their level-$k$ labels, and run cluster validation metrics to assess the compactness of the emerging point clouds. We report the results in Table~\ref{tab:clustering}. The selected cluster evaluation metrics are the Silhouette score, (a standard choice), 
the Calinski-Harabasz score, and S-Dbw, the best metric according to a comparative review by Liu et al.~\cite{clustering}

Our results show that our models systematically outperform XENT at every hierarchical level in terms of cluster validation metrics. Furthermore, HT-AL provides better S-Dbw scores than HAF. This confirms our expectation: S-AL allows for a structured internal representation learning, so that ontologically-similar labels are projected into contiguous areas of the feature space.
\subsection{Explainability}
Finally, we inspect the heatmaps produced by XAI algorithms, looking for correlations between ontological label proximity and generated heatmap similarity. We generate 48 million distance measures (10000 images in the CIFAR100 test set $\times$ 100 possible classes $\times$ 4 selected models $\times$ 3 XAI algorithms $\times$ 4 distance metrics), and partition them according to the original two-levels CIFAR100 taxonomy or the more relevant five-levels one, thus approaching the 100M data points. For space constraints, below we only show a selected subset of results.
\begin{figure}[t!]
\centering
\includegraphics[width=\textwidth]{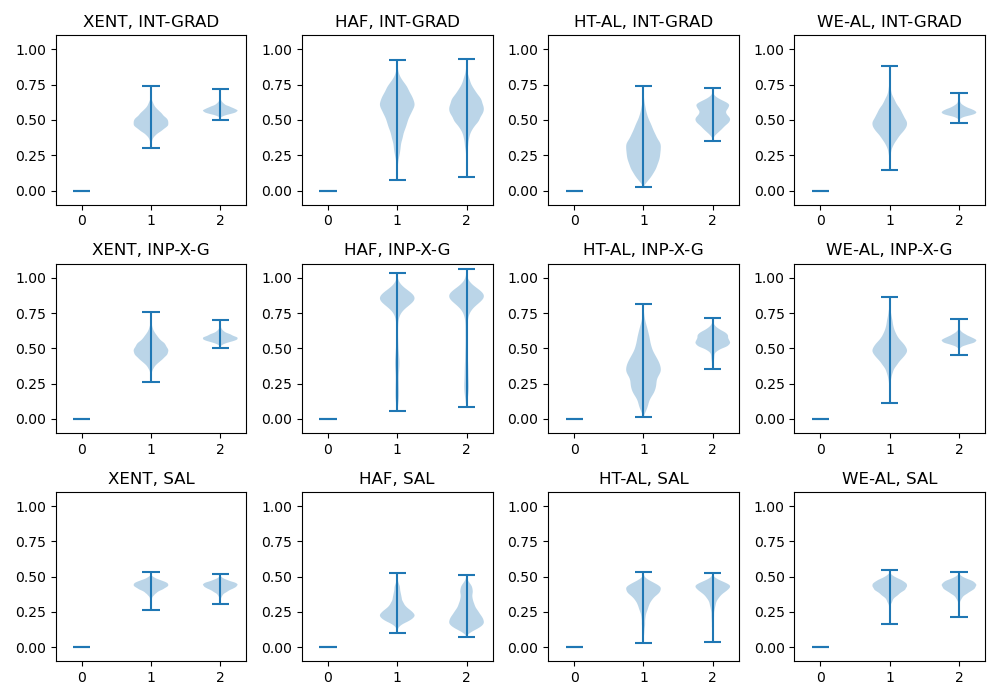}
\caption{Relation between ontological label distance and produced heatmaps}
\label{fig:v1}
\end{figure}
For Figure~\ref{fig:v1}, we focused on the \textit{Progressive Binarisation} metric, and we partitioned on the original two-levels CIFAR100 taxonomy - which is why the x-axis has violins for the semantic distances of 0, 1, and 2. Every row of panels corresponds to an explainer (respectively, \textit{Integrated Gradients}, \textit{Input X Gradients} and \textit{Saliency}), and every column corresponds to a selected model: XENT, HAF, HT-AL, WE-AL. 
For all panels, a violin plot at position $k$ represents the aggregation of heatmap distances that involved labels at ontological distance $k$. For the \textit{husky-flute} example image, the \textit{husky} label has distance 0 (true label), the \textit{dog} label has distance 1 (similar label), and the \textit{flute} label has distance 2 (far-away concept). Thus, the \textit{husky-heatmap} / \textit{dog-heatmap} distance would end up in the 2-distance violin (regardless of model, metric, and explainer). We note that all distance-0 violins correspond to trivial same-heatmap comparisons, and therefore correspond to 0. Our goal is to have higher/wider violins for distance-2, with respect to distance-1. We observe that XENT has a non-null baseline, and this is likely due to the visual similarity of images belonging to similar classes. HAF produces almost identical violins for distance 1 and distance 2 in all panels. Conversely, S-AL models produce pairs of distinguishable violins - especially for the HT-AL model: this shows that the auxiliary information injected in the augmented labels positively conditioned the model's training.
Finally, we note how \textit{Saliency} systematically fails to detect any ontology-driven distance.

\begin{figure}[t!]
\centering
\includegraphics[width=\textwidth]{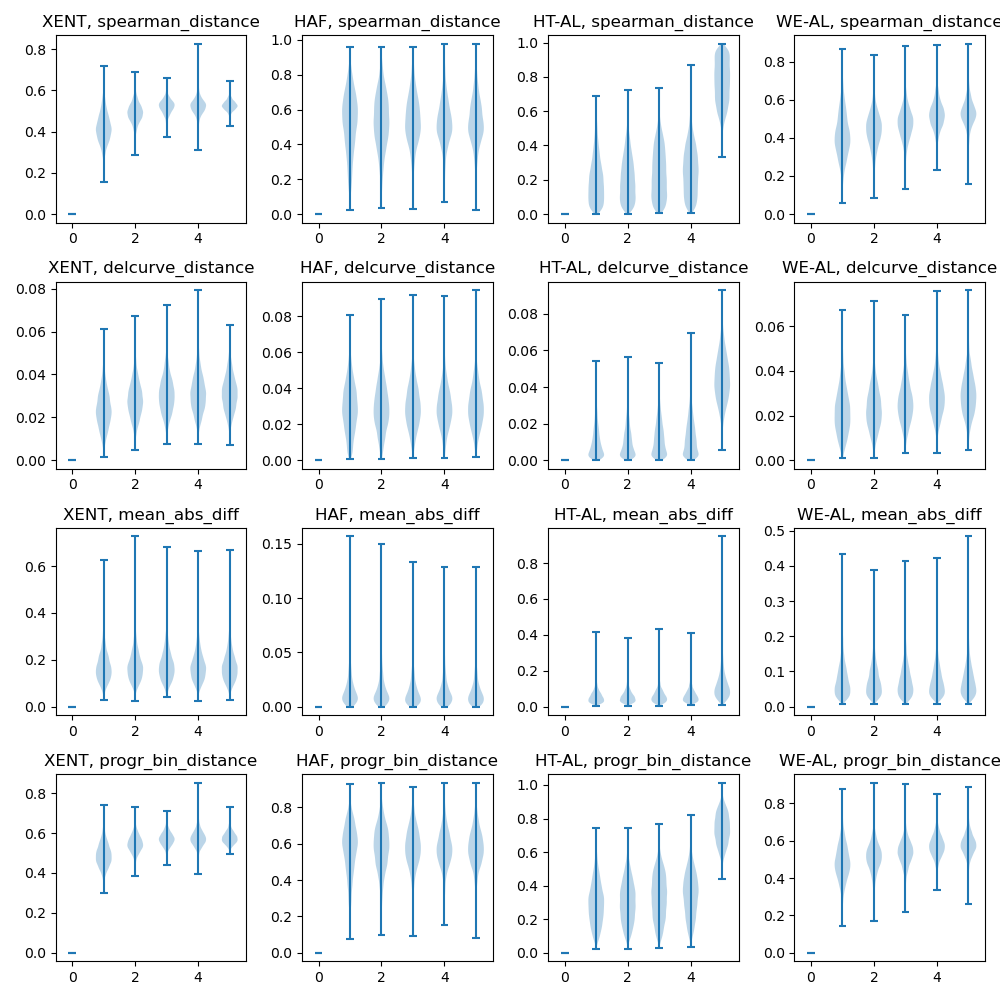}
\caption{Relation between ontological label distance and produced heatmaps}
\label{fig:v2}
\end{figure}
For Figure~\ref{fig:v2}, we used the full 5-levels CIFAR100 ontology, all models (one per column), all distance metrics (one per row), and we focused on the \textit{Integrated Gradients} explainer. As for the previous case, we observe how XENT has a non-trivial baseline, HAF does not capture the taxonomical hierarchy information, while HT-AL does. Curiously, the heatmaps corresponding to distance-5 labels seem remarkably more different than the others, while only minor differences appear from distance-1 to distance-4 labels. This seems to correlate with the two blocks visible in Figure~\ref{fig:feat_dist}, second panel.
\section{Conclusions and Future Work}
\label{sec:end}
In this paper, we introduced \textbf{Semantically-Augmented Labels}, a general approach to combine OHE labels with arbitrary auxiliary semantic information so that the resulting augmented ground truth can be used for image classification training procedures, without the need for custom loss functions or model architectures. Starting from the benchmark dataset of CIFAR100, we showed how to apply our approach to ontological information (\textbf{HT-AL}) and GloVe-derived word embedding vectors (\textbf{WE-AL}). We conducted experiments and analysed the impact and implications of our approach in terms of machine learning performance, organisation of the learned feature space, and characterisation of the produced explanation heatmaps. We showed how our approach allows to train ML models whose accuracy is competitive with respect to a classically trained baseline; at the same time, our models showed interesting results in terms of generalisation to super-classes (level$>$1 error rates), organisation of the feature space (cluster quality) and differentiation between explanations (heatmap distances). 

We hope this approach can provide a useful middle ground in the debate between post-hoc explainability and the design of custom interpretable models. With the former, we share the goal of exploiting as much as possible existing standard architectures and loss functions, since (i) they have proved to work very well and (ii) this allows to tap into a plethora of boilerplate code and existing repositories. On the other hand, we share with the latter the intuition that intervening in the learning phase, rather than after it, allows for more room for action.

Concerning directions for future work, the first natural evolution is to tackle bigger datasets, such ad iNaturalist or ImageNet. While both image datasets are paired with hierarchical taxonomies of labels, we are interested in exploring the impact of injecting different types of embedding-based semantic information. 

Another line of research we are already exploring is whether S-AL models can display increased robustness to adversarial attacks. Intuitively, it is often speculated that adversarial attacks exploit anomalies 
in the decision boundary of an image classifier. 
The injection of auxiliary semantic information could allow for more control of the learned feature space, thus making it more adherent to human intuition - for instance, projecting \textit{turtle} and \textit{rifle} images on non-neighbouring areas. This
might make the decision boundary smoother and less prone to human-counterintuitive mistakes.


\section*{Acknowledgments}
This work has been partially supported by  the
 “National Centre for HPC, Big Data and Quantum
Computing” - Spoke 1 funded by NextGenerationEU.


\bibliographystyle{abbrv}
\bibliography{biblio}

\end{document}